\documentclass{article}
\usepackage{spconf,amsmath,graphicx}
\usepackage{comment}
\title{Towards End-to-End Code-Switching Speech Recognition}
\name{Ne Luo\textsuperscript{*}\thanks{\textsuperscript{*}The joint contributors}, Dongwei Jiang\textsuperscript{*}, Shuaijiang Zhao\textsuperscript{*}, Caixia Gong\textsuperscript{*},  Wei Zou, Xiangang Li}

\address{
  AI Labs, Didi Chuxing, Beijing, China \\
  \{luone\_i, jiangdongwei, zhaoshuaijiang, gongcaixia, zouwei, lixiangang\}@didiglobal.com
  }

\ninept

\begin{document}

\maketitle

\begin{abstract}

Code-switching speech recognition has attracted an increasing interest recently, but the need for expert linguistic knowledge has always been a big issue.
End-to-end automatic speech recognition (ASR) simplifies the building of ASR systems considerably by predicting graphemes or characters directly from acoustic input. 
In the mean time, the need of expert linguistic knowledge is also eliminated, which makes it an attractive choice for code-switching ASR.
This paper presents a hybrid CTC-Attention based end-to-end Mandarin-English code-switching (CS) speech recognition system and studies the effect of hybrid CTC-Attention based models, different modeling units, the inclusion of language identification and different decoding strategies on the task of code-switching ASR. 
On the SEAME corpus, our system achieves a mixed error rate (MER) of 34.24\%.

\end{abstract}
\section{Introduction}
As multilingual phenomenon becoming more and more common in real life \cite{baker2011foundations}, there has been an increasing interest in code-switching speech recognition.
Code-switching speech is defined as speech which contains more than one language within an utterance \cite{auer2013code}.

Several challenges appear in this area, including the lack of training data for language modeling \cite{adel2015syntactic}, the co-articulation effects \cite{vu2012first}, and the need of expert linguistic knowledge.
Therefore, it is difficult to build a good ASR system that can handle code-switching phenomenon. 
Previous work mainly focus on the first two challenges. 
Statistical machine translation (SMT) is used to generate artificial code-switching texts \cite{vu2012first}. Recurrent neural network language models (RNNLMs) and factored language models (FLMs) with integration of part-of-speech (POS) tag, language information, or syntactic and semantic features are proposed to improve the performance of language modeling to code-switching speech \cite{adel2015syntactic,adel2013combination,adel2014features}.
To tackle the co-articulation problem, speaker adaptation, phone sharing and phone merging are applied \cite{vu2012first}. Additionally, language information is incorporated into ASR systems by introducing a language identifier \cite{bhuvanagiri2010approach,DBLP:conf/icassp/LyuLCH06,weiner2012integration}. 

Recently, end-to-end speech recognition systems \cite{amodei2016deep} are becoming increasingly popular while achieving promising results on various ASR benchmarks.
End-to-end systems reduce the effort of building ASR systems considerably by predicting graphemes or characters directly from acoustic information without predefined alignment.

In a code-switching scenario, we believe end-to-end models have a competitive advantage over traditional systems since we do not need expert linguistic knowledge on the objective languages and the burden of generating specific lexicons can be relieved. 
There is only one previous work in building end-to-end code-switching speech recognition systems \cite{DBLP:conf/icassp/SekiWHRH18}, but they use artificially generated data, producing by concatenating monolingual utterances, rather than spontaneous code-switching speech.

Two major types of end-to-end architectures are: the connectionist temporal classification (CTC) \cite{graves2006connectionist,kim2017joint} and attention-based method \cite{prabhavalkar2017comparison,chorowski2015attention,chan2016online}.
CTC objective can be used to train end-to-end systems that directly predict grapheme sequences without requiring a frame-level alignment of the target labels for a training utterance.
Attention-based method consists of an encoder network and an attention-based decoder, which maps acoustic speech into high-level representation and recognizes symbols conditioned on previous predicts, respectively.
\cite{kim2017joint} presents a joint CTC-Attention multi-task learning model that combines the benefit of both two types of systems. Their model achieves state-of-the-art results on multiple public benchmarks while improving the robustness and speed of convergence compared to other end-to-end models.

In this work, we apply a framework similar to the joint CTC-Attention model on Mandarin-English code-switching speech to observe whether it can match the performance of traditional systems while preserving the benefits of end-to-end models. We also study the effect of different modeling units, the inclusion of language identification and different decoding strategies on end-to-end code-switching ASR. All of our experiments are conducted on the SEAME corpus \cite{Lyu2010SEAMEAM}.

The rest of this paper is organized as follows. 
Section \ref{section_cs} introduces the details of attention and CTC framework. 
The end-to-end based code-switching speech recognition, including modeling units, language identification and decoding strategies are studied in Section \ref{section_strategy}. 
Section \ref{section_exp} describes the details of our model and analyzes the results of our experiments.
Section \ref{section_conclusion} draws some conclusions and discusses our future work.

\section{End-to-end framework} \label{section_cs}

\subsection{Connectionist temporal classification (CTC)}
Key to CTC \cite{graves2006connectionist} is that it removes the need of prior alignment between input and output sequences. Taking the network outputs as a probability distribution over all possible label sequences, conditioned on a given input sequence $ x $, we can define an objective function to maximize the probabilities of the correct labelling.
To achieve this, an extra ‘blank’ label denoted $\langle b \rangle$ is introduced to map frames and labels to the same length, which can be interpreted as no target label.
CTC computes the conditional probability by marginalizing all possible alignments and assuming conditional independence between output predictions at different time steps given aligned inputs.

Given a label sequence $y$ corresponding to the utterance $x$, where $y$ is typically much shorter than the $x$ in speech recognition.
Let $\beta(y, x)$ be the set of all sequences consisting of the labels in $\mathcal{Y} \cup {\langle b \rangle}$, which are of length $|x| = T$, and which are identical to $y$ after first collapsing consecutive repeated targets and then removing any blank symbols (e.g., $A{\langle b\rangle}AA{\langle b \rangle}B \to AAB$).
CTC defines the probability of the label sequence conditioned on the acoustics as Equation \ref{CTC_prob}.

\begin{equation}
  P_{CTC}(y|x) = \sum_{\hat{y} \in \beta(y,x)} P(\hat{y}|x)
                               = \sum_{\hat{y}\in\beta(y,x)} \prod_{t=1}^{T} P(\hat{y}_t|x) \label{CTC_prob}
\end{equation}

\subsection{Attention based models}
Chan et al. \cite{LAS} proposed Listen, Attend and Spell (LAS), a kind of neural network that learns to transcribe speech utterances to characters.
LAS is based on the sequence to sequence learning framework with attention and consists of two sub-modules: the listener and the speller.

Most attention models used in speech recognition share similar structure as LAS and is often used to deal with variable length input and output sequences.
An attention-based model contains an encoder network and an attention based decoder network.
The attention-based encoder-decoder network can be defined as:
\begin{equation}
  h = Encoder(x)  \label{encoder},
\end{equation}
\begin{equation}
  P(y_t|x,y_{1:t-1})= AttentionDecoder(h, y_{1:t-1}) \label{decoder},
\end{equation}
where $Encoder(\cdot)$ can be Long Short-Term Memory (LSTM) or Bidirectional LSTM (BLSTM) and $AttentionDecoder(\cdot)$ can be LSTM or Gated Recurrent Unit (GRU).

The encoder network maps the input acoustics into a higher-level representation $h$.
The attention based decoder network predicts the next output symbol conditioned on the full sequence of previous predictions and acoustics, which can be defined as $P(y_t|x,y_{1:t-1})$.

The attention mechanism selects (or weights) the input frames to generate the next output element. Two of the main attention mechanisms are: content-based attention \cite{bahdanau2014neural} and the location-based attention \cite{chorowski2015attention}. Borrowed from neural machine translation, content-based attention can be directly used in speech recognition. 
For location-based attention, location-awareness is added to the attention mechanism to better fit the speech recognition task.

\section{Methods}\label{section_strategy}

\subsection{Hybrid CTC-Attention based models} \label{subsec_mtl}

Inspired by \cite{kim2017joint}, we add a CTC objective function as an auxiliary task to train the encoder of attention model. The forward-backward algorithm of CTC enforces a monotonic alignment between input and output sequences, which helps the attention model to converge. 
The attention decoder learns label dependency, thus often shows improved performance over CTC when no external language model is used.

We combine CTC and attention model by defining a hybrid CTC-Attention objective function utilizing two losses:
\begin{equation}
  L_{MTL} = \lambda L_{Att} + (1-\lambda)L_{CTC},
\end{equation}
where $ \lambda $ is a tunable hyper-parameter in the range of [0, 1], dictating the weight assigned to attention loss.

\subsection{Acoustic modeling units} \label{subsec_units}
Syllable and character are common acoustic modeling units for Mandarin speech recognition system. We choose character as Mandarin acoustic modeling unit as it is the most common choice for end-to-end Mandarin ASR and it has shown state-of-the-art performance on several public benchmarks \cite{zou2018comparable,zhou2018comparison}. As for English, the frequently used acoustic modeling units in end-to-end speech recognition systems are character \cite{kim2017joint,LAS} and subword \cite{rao2017exploring,zenkel2017subword}. In this paper, we explore two acoustic modeling units combination for Mandarin and English code-switching speech recognition: character units for both languages (Character-Character), and character units for Mandarin plus subwords units for English (Character-Subword).

Character-Character model takes acoustic features as input and outputs sequences consisting of Chinese and English characters. Let $Y$ be the output sequences, $Y = (\langle sos \rangle, y_1, y_2, .., y_T, \langle eos \rangle)$, $y_i \in \{y_{CH}, y_{EN}, \langle apostrophe \rangle, \langle space \rangle, \langle unk \rangle\}$,  where $ y_{CH} $ contains a few thousand frequently used Chinese characters, $ y_{EN} $ contains 26 English characters, and $\langle sos \rangle, \langle eos \rangle $ represents the start and the end of a sentence respectively.

Character-Subword model is built with a vocabulary containing Chinese characters and English subwords. In this paper, we adopt Byte Pair Encoding (BPE) \cite{sennrich2015neural} as subword segmentation method. BPE is an algorithm that originally used in data compression, it replaces the most frequent pair of bytes (characters) in a sequence with a single and unused byte (character sequence). We iteratively replace the most frequent pair of symbols with a new symbol, and every new symbol is added to the subword set. The process ends when the amount of subword reaches the value we set. We insert a special symbol '\_' before every English word to represent the start of words. By the time the subword set is generated, we splits English words into subwords by greedily segmenting the longest subword in a word. After decoding, words sequences are reconstructed from subword-based output sequences by replacing all the word boundary marks in subwords with spaces.

\begin{figure}[th]
  \centering
  \includegraphics[width=\linewidth,trim=0 166 0 166,clip]{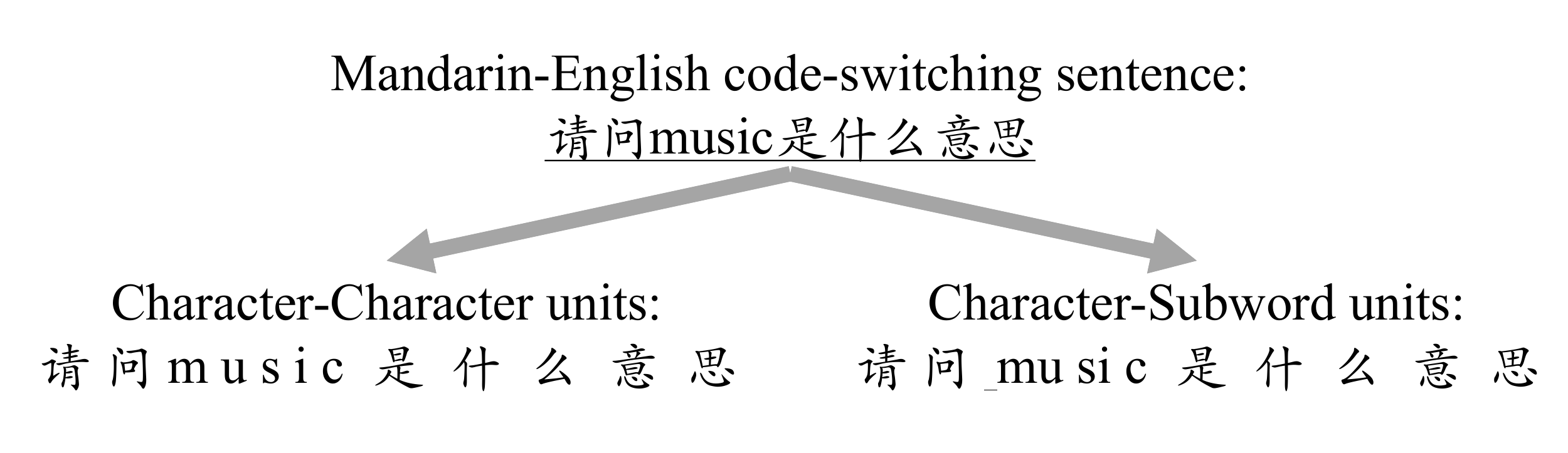}
  \caption{An example of converting one Mandarin-English code-switching sentence into two kinds of modeling units.}
  \label{fig:chinglish_example}
\end{figure}

According to the segmentation methods above, a Mandarin-English code-switching sentence can be converted into two kinds of modeling units, which is shown in Fig.~\ref{fig:chinglish_example}.

\subsection{Joint language identification (LID)} \label{subsec_units}
In code-switching ASR, words with similar pronunciation from different languages are very likely to be recognized incorrectly. To deal with this problem, we consider to include language identification in our system. Specifically, we propose two strategies to incorporate LID into our system. 

One is LID-Label, which is similar to Seki el al.'s work \cite{DBLP:conf/icassp/SekiWHRH18}. In this strategy, we use an augmented vocabulary, adding LID 'CH' and 'EN' as part of output symbols. The decoder network predicts corresponding LID before the following characters/subwords once it meets code-switching points. In this way the network is forced to learn language information. 

The other method is training networks to recognize speech and language simultaneously through multi-task learning framework, LID-MTL for short. Similar to \cite{weiner2012integration}, we create the alignment result of Chinese characters and English words in advance and generate LID sequences based on the alignments. Then we add in a new network that shares the encoder with attention model and CTC model. The loss of this new network is cross entropy of predicted LID and ground truth LID from alignments. We combine their losses using Equation \ref{equ:mtl_lid}:
\begin{equation}\label{equ:mtl_lid}
  L_{MTL} = \lambda_{Att} L_{Att} + \lambda_{CTC}L_{CTC} + \lambda_{LID} L_{LID},
\end{equation}
where $ \lambda_{Att} $, $\lambda_{CTC} $ and $ \lambda_{LID} $ are tunable hyper-parameters with a sum of 1, dictating the weight assigned to the corresponding loss.

\begin{figure}[th]
  \centering
  \includegraphics[width=\linewidth,trim=200 50 200 200,clip]{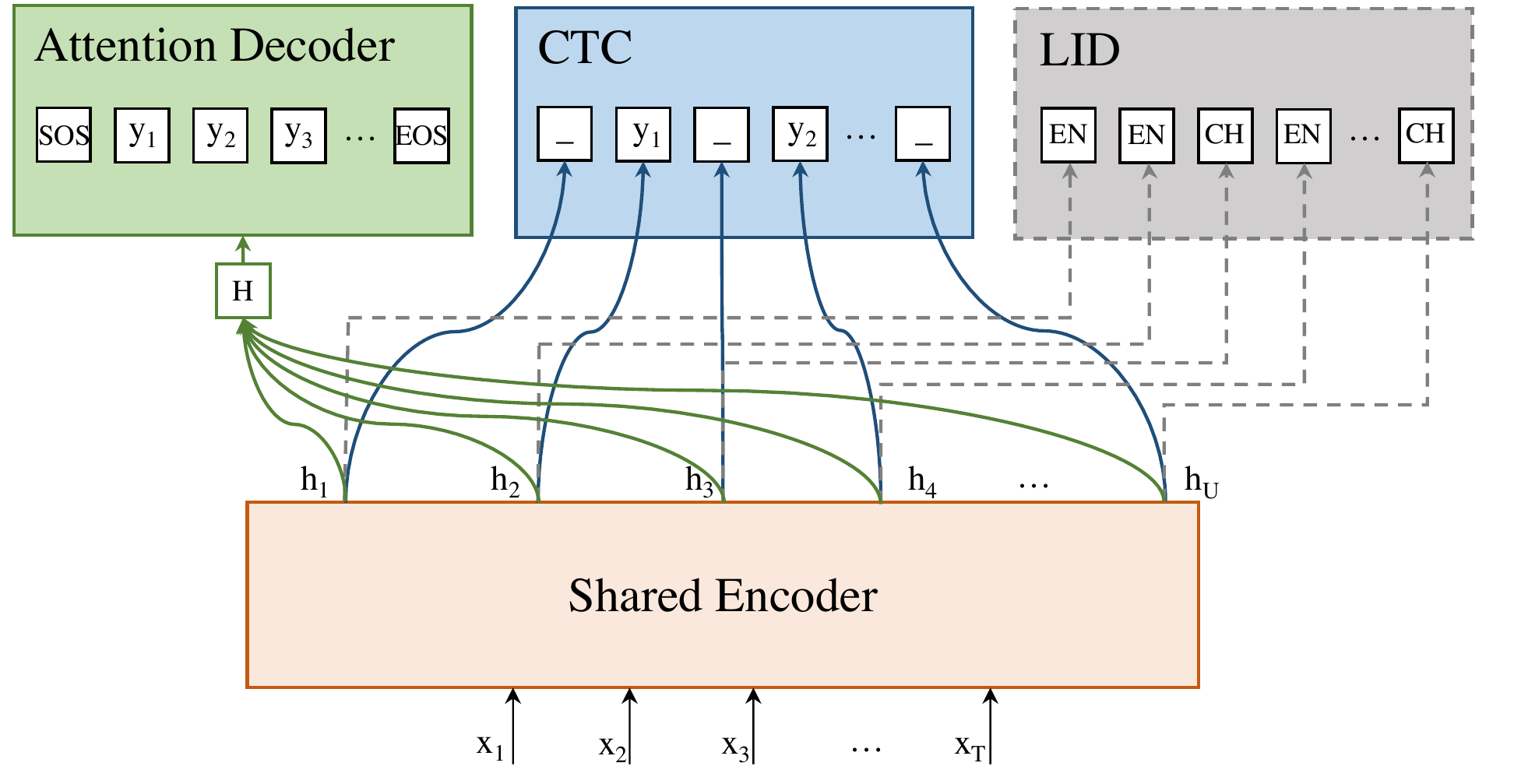}
  \caption{Our proposed joint speech recognition and language identification multi-task learning framework. The encoder is shared by Attention decoder, CTC and LID component. It transforms input sequence $ x $ into high level features $ h $. Attention decoder and CTC generate output sequence $ y $, while LID component outputs language IDs for each frame.}
  \label{fig:lid}
\end{figure}

Fig.~\ref{fig:lid} shows the architecture of proposed LID-MTL model.

\subsection{Decoding Strategy} \label{subsec_bs}
As we are using hybrid CTC-Attention model, the joint CTC-Attention beam-search decoding introduced in \cite{Hori2017JointCD} is applied as our basic strategy.


After some experiments and analysis of decoding results, we find an interesting phenomenon: although the TER of our end-to-end model is relative low, the MER is higher than our expectation because some of the final winners in beam search contain subword sequences that cannot form valid words. In order to overcome this problem, we generated a word dictionary containing a few thousand frequently used English words and words appeared in the SEAME train set. And we also developed two decoding strategies to try and increase the odds of candidates that can form correct words being selected.
\begin{itemize}
\item \textbf{Decode1}: At the end of beam search, we only choose candidates whose subword sequences form correct words to compete for a final winner.

\item \textbf{Decode2}: During beam search, we discard candidates whose subword sequence cannot form correct word.
\end{itemize}
\section{Experiments}\label{section_exp}

   \subsection{Data}
   We conduct our experiments on the SEAME (South East Asia Mandarin-English) corpus. SEAME is a 66.8 hours Mandarin-English code-switching corpus containing spontaneous conversation and interview talks recorded from Singapore and Malaysia speakers. The corpus includes 155 speakers, where 115 in them are Singaporean and the rest are Malaysian. The ratio of gender is quite balanced, in which female and male accounts for 55\% and 45\% respectively. There is a small proportion of monolingual segments in this corpus, only 12\% and 6\% of the transcribed segments are Mandarin and English monolingual utterances respectively. We divide the SEAME corpus into three sets (train, development and test) based on several criteria like gender, speaking style, speaker nationality and so on. The detailed statistics of the SEAME corpus are presented in Table \ref{tab:seame}.

\begin{table}[!h]
\centering
\caption{Statistics of the SEAME corpus}
\begin{tabular}{lcccc}
\hline\noalign{\smallskip}
 & Train set & Dev set & Test set & Total \\
\noalign{\smallskip}\hline\noalign{\smallskip}
Speakers & 137 & 9 & 9 & 155 \\
Duration (hours) & 59.4 & 4.4 & 3.0 & 66.8 \\
Utterances & 47966 & 4328 & 2333 & 54627 \\
\noalign{\smallskip}\hline
\end{tabular}
\label{tab:seame}
\end{table}

  \subsection{Training}
      The model we use is a hybrid CTC-Attention model.
      The shared encoder has 2 convolutional layers, followed by 4 bi-directional GRU layers with 256 GRU units per-direction, interleaved with 2 time-pooling layers which results in an 4-fold reduction of the input sequence length. 
      The decoder model has 1 GRU layer with 256 GRU units and output consists of 2376 Chinese characters, 1 unknown character, 1 sentence start token, 1 sentence end token, and the English character/subword set.

      During training stage, scheduled sampling and unigram label smoothing are applied as described in \cite{chiu2017state,bahdanau2016end,chorowski2016towards}. Adam optimization method with gradient clipping is used for optimization. We initialize all the weights randomly from an isotropic Gaussian distribution with variance 0.1 and learning rate is decayed from 5e-4 to 5e-5 during training. The model is trained using TensorFlow \cite{abadi2016tensorflow}.
      
      A recurrent neural network based language model (RNNLM) is incorporated into the hybrid CTC-Attention based model.
      The RNNLM is composed of 2 LSTM layers of 800 hidden units each. It has the same output vocabulary as the hybrid CTC-Attention based model. 
      The RNNLM is trained with the SEAME train set and validated on the dev set. 
      The AdaDelta algorithm with gradient clipping is used for the optimization with an initial learning rate of 0.05.
      All experiments we conduct below incorporate with RNNLM.

  \subsection{Choice of MTL weight}
      We first conduct experiments using different choice of MTL weight with Character-Character model. As shown in Table \ref{tab:mtl}, our model get lowest MER with $ \lambda = 0.8 $. This is consistent with our expectation that models trained with multi-task objective function perform better than using attention objective. Therefore, we choose $ \lambda = 0.8 $ in the following experiments.

\begin{table}[!h]
\centering
\caption{MERs (\%) of different hyper-parameter $ \lambda $ on the development set (Dev) and test set (Test) of SEAME for character based systems.}
\begin{tabular}{cccccc}
\hline\noalign{\smallskip}
$ \lambda $ & Dev & Test \\
\noalign{\smallskip}\hline\noalign{\smallskip}
0.2 & 39.72 & 40.94 \\
0.5 & 38.24 & 39.97 \\
0.8 & 37.59 & \textbf{39.31} \\
1.0 & 38.03 & 40.27 \\
\noalign{\smallskip}\hline
\end{tabular}
\label{tab:mtl}
\end{table}

  \subsection{Effect of using different modeling units}
      The subword set we use is trained on English segments of the SEAME train set using SentencePiece \cite{Kudo2018SentencePieceAS}. We generate two sets of subwords (200 and 500 respectively) to observe the effect of number of subwords on final result.

      As shown in Table \ref{tab:units}, Character-Character models perform worse than both Character-Subword models. We believe it is because Character-Subword model retains an reasonably similar correspondence between output unit and audio segment length, which could be a problem with simply using character for English.
      Subword 500 model also performs worse than subword 200 model, due to the sparser distribution of subword on this small corpus.

\begin{table}[!h]
\centering
\caption{MERs (\%) on the development set (Dev) and test set (Test) of SEAME. Mixed output units consists of character and subword, and 'mixed-200' means having 200 subwords, 'mixed-500' means having 500 subwords.}
\begin{tabular}{ccccc}
\hline\noalign{\smallskip}
Model & Output units  & Dev & Test\\
\noalign{\smallskip}\hline\noalign{\smallskip}
Att + CTC & character & 37.59 & 39.31 \\
Att + CTC & mixed-200 & 35.44 & \textbf{37.83} \\
Att + CTC & mixed-500 & 36.33 & 38.05 \\
\noalign{\smallskip}\hline
\end{tabular}
\label{tab:units}
\end{table}

  \subsection{Joint language identification}
     
      The results of LID-Label and LID-MTL with different LID weights are listed in Table \ref{tab:lid}. Unfortunately, we do not find much improvement of LID-Label over baseline. We suspect it is because the frequent language switching in SEAME makes it harder for the model to accurately predict language ID.
      
      On the other hand, LID-MTL gives significant improvement over the baseline, essentially backing up the theory that adding a third objective function on shared encoder would be beneficial to code-switching system.
      When we set LID weight to be larger than 0.1, performance of the model becomes much worse. So from now on, we would use 0.1 throughout our experiments.
  
\begin{table}[!h]
\centering
\caption{MERs (\%) on the development set (Dev) and test set (Test) of SEAME. $ \lambda_{LID} $ in the table represents the weight of LID loss in LID-MTL, while $ \lambda_{Att} = 0.8 $,  $ \lambda_{CTC} = 0.2 - \lambda_{LID}$.}
\begin{tabular}{cccc}
\hline\noalign{\smallskip}
Model & $\lambda_{LID}$ & Dev & Test \\
\noalign{\smallskip}\hline\noalign{\smallskip}
Att + CTC   & - & 35.44 & 37.83 \\
LID-Label & -  & 35.48 & 37.98 \\
LID-MTL & 0.05 & 34.45 & 37.03 \\
LID-MTL & 0.10  & 34.13 & \textbf{36.48} \\
LID-MTL & 0.20  & 35.43 & 37.82 \\
\noalign{\smallskip}\hline
\end{tabular}
\label{tab:lid}
\end{table}

  \subsection{Effect of different decoding strategy}
      Table \ref{tab:decode} shows MERs on SEAME using different decoding strategies. It is obvious that Decode2 imposes a stronger restriction on beam search candidates, but it may also remove correct decoding results because some mistakes occur in early stage of decoding. However, the final MER of Decode2 is lower than that of Decode1. This seems to suggest our end-to-end model is having a hard time to relate different parts of subword together and one possible explanation is that the size of SEAME is too small. We are interested to figure out how the performance of Decode1 and Decode2 would be on bigger Mandarin-English code-switching corpora. 
          
\begin{table}[!h]
\centering
\caption{MERs (\%) on development set (Dev) and test set (Test) of SEAME with different decoding strategies for output units are mixed-200.}
\begin{tabular}{cccc}
\hline\noalign{\smallskip}
Model & Decoding Strategy & Dev & Test\\
\noalign{\smallskip}\hline\noalign{\smallskip}
Att + CTC & Beam search & 35.44 & 37.83 \\
Att + CTC & Decode1 & 35.02 & 37.35  \\
Att + CTC & Decode2 & 34.67 & 36.59   \\
LID-MTL & Beam search & 34.13 & 36.48 \\
LID-MTL & Decode1 & 32.97 & 35.31 \\
LID-MTL & Decode2 & 32.31 & \textbf{34.24} \\
\noalign{\smallskip}\hline
\end{tabular}
\label{tab:decode}
\end{table}

\section{Conclusions}\label{section_conclusion}
  In this work, we created a hybrid CTC-Attention based end-to-end Mandarin-English code-switching speech recognition system that outperforms most traditional code-switching speech recognition systems on the SEAME corpus.
  We also studied the effect of different modeling units, inclusion of language identification and different decoding strategies. 
  As for acoustic modeling units, we found the combination of Chinese character and English subword to be the optimal choice. 
  When it comes to the inclusion of LID, we managed to get significant reduction in MER by adding a third cross entropy objective on LID.
  Last but not least, a novel decoding strategy that considers word information during beam search would also help to reduce the MER of code-switching speech recognition.
  
  In the future, we would like to conduct a more thorough investigation on our strategies and explore how our proposed framework performs on other code-switching tasks (possibly with more than two languages).


\bibliographystyle{IEEEbib}
\bibliography{cs_asr}

\end{document}